\documentclass[letterpaper, 10 pt, conference]{ieeeconf}  

\IEEEoverridecommandlockouts                              

\usepackage{cite}
\usepackage{graphicx}
\usepackage{multirow}
\usepackage{booktabs}

\title{\LARGE \bf
Robust Imitation Learning for Mobile Manipulator\\Focusing on Task-Related Viewpoints and Regions
}

\author{Yutaro Ishida$^{1}$, Yuki Noguchi$^{1}$, Takayuki Kanai$^{1}$, Kazuhiro Shintani$^{1}$ and Hiroshi Bito$^{1}$
\thanks{$^{1}$Toyota~Motor~Corporation, 1 Toyota-Cho, Toyota~City, Aichi 471-8571, Japan
        {\tt\small \{yutaro\_ishida, yuki\_noguchi\_ad, takayuki\_kanai, kazuhiro\_shintani, hiroshi\_bito\}@mail.toyota.co.jp}}%
}

\begin{document}

\bstctlcite{IEEEexample:BSTcontrol}

\maketitle
\thispagestyle{empty}
\pagestyle{empty}

\begin{abstract}

We study how to generalize the visuomotor policy of a mobile manipulator from the perspective of visual observations.
The mobile manipulator is prone to occlusion owing to its own body when only a single viewpoint is employed and a significant domain shift when deployed in diverse situations.
However, to the best of the authors’ knowledge, no study has been able to solve occlusion and domain shift simultaneously and propose a robust policy.
In this paper, we propose a robust imitation learning method for mobile manipulators that focuses on task-related viewpoints and their spatial regions when observing multiple viewpoints.
The multiple viewpoint policy includes attention mechanism, which is learned with an augmented dataset, and brings optimal viewpoints and robust visual embedding against occlusion and domain shift.
Comparison of our results for different tasks and environments with those of previous studies revealed that our proposed method improves the success rate by up to 29.3 points.
We also conduct ablation studies using our proposed method.
Learning task-related viewpoints from the multiple viewpoints dataset increases robustness to occlusion than using a uniquely defined viewpoint.
Focusing on task-related regions contributes to up to a 33.3-point improvement in the success rate against domain shift.
\end{abstract}

\section{Introduction}
\label{sec:intoroduction}


In recent years, mobile manipulators (MMs) have entered the spotlight owing to their ability to perform various tasks in various locations, which can be attributed to their high mobility  \cite{cite:hsr_toyota_iros, cite:pr2, cite:fetch, cite:tiago}.
Extensive research and development have been conducted to explore their societal implementation, including recognition \cite{cite:hsr_ishida_dataset, cite:hsr_ishida_accelerator, cite:hsr_ono}, control \cite{cite:hsr_toyota_robomech}, autonomy \cite{cite:hsr_autonomy}, simulation \cite{cite:hsr_matsusaka} and robotics competitions \cite{cite:robocup, cite:hsr_luis}.
Researchers are exploring robot learning techniques such as imitation learning and reinforcement learning to develop learned policies that can achieve more advanced tasks than traditional hand-crafted programming \cite{cite:bcz, cite:rt1, cite:aloha, cite:mobile_aloha}.

In robot learning, a method called visuomotor policy is commonly used to predict the next action based on visual observation.
However, there are two problems that are typically encountered when using the policy on MMs.

\begin{figure}[t]
    \begin{center}
        \includegraphics[width=1.0\linewidth]{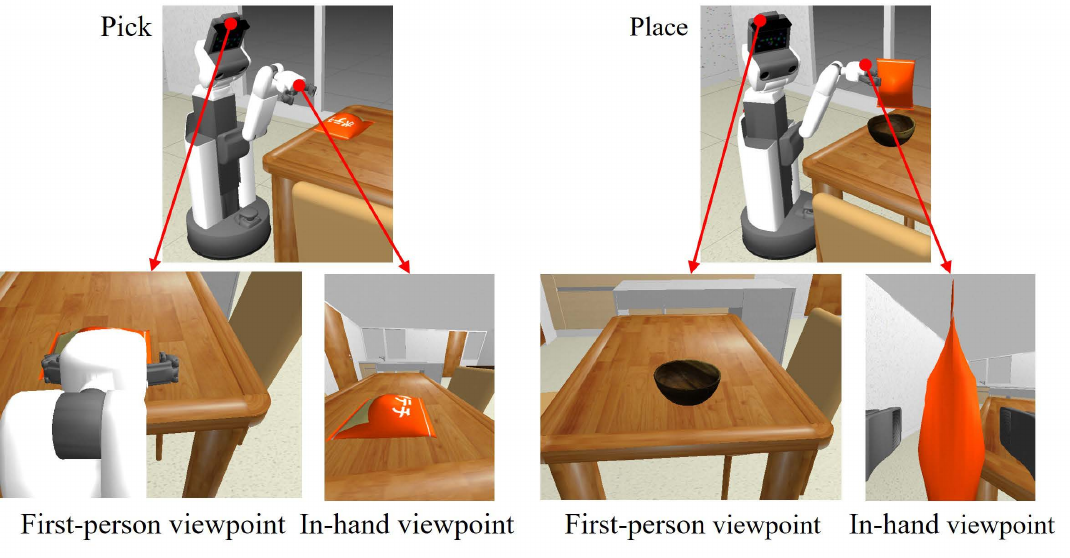}
        \caption{Example of occlusion of internal viewpoints on the mobile manipulators. Left: first-person viewpoint is occluded by the body of the MM in pick task. Right: in-hand viewpoint is occluded by the grasped object in place task.}
        \label{fig:explanation_occlusion}
        \vspace{-2mm}
    \end{center}
\end{figure}

\begin{figure}[t]
    \begin{center}
        \includegraphics[width=1.0\linewidth]{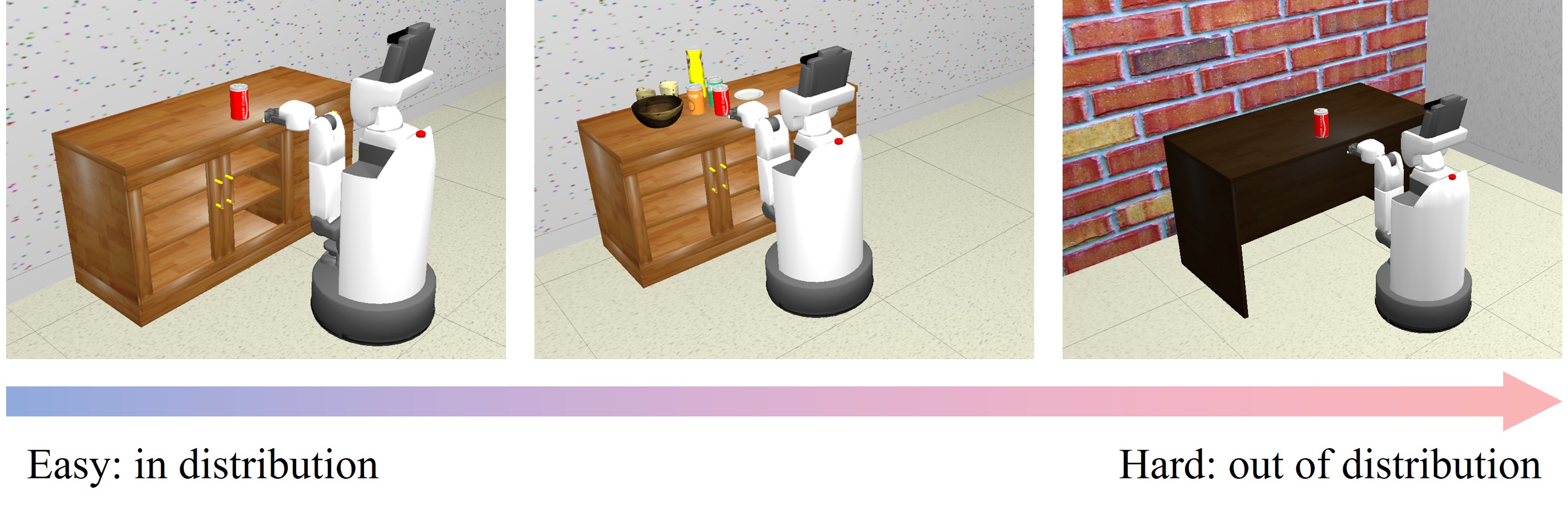}
        \caption{Example of visual observation domain shift. Left: environments for training the policy. Middle: distractor objects cause the minor change. Right: unknown furniture causes the major change.}
        \label{fig:explanation_ood}
        \vspace{-2mm}
    \end{center}
\end{figure}



First, MMs are prone to occlusion in visual observation when only a single internal viewpoint is used.
For instance, in the pick task shown in Fig. \ref{fig:explanation_occlusion} left, the first-person viewpoint can easily be occluded by the body of the MM.
In contrast, in the place task shown in Fig. \ref{fig:explanation_occlusion} right, the in-hand viewpoint can easily be occluded by the grasped object.
To address this, the policy needs to be able to handle occlusion by focusing on task-related viewpoints from multiple internal viewpoints depending on the situation: using in-hand viewpoint for pick task (Fig. \ref{fig:explanation_occlusion} left) and first-person viewpoint for place task (Fig. \ref{fig:explanation_occlusion} right).

Second, compared with a manipulator without a mobile base, MMs are more prone to significant domain shifts in visual observation owing to their varied usage \cite{cite:rt1, cite:rt2}.
For example, in the pick-red-can task, training environments (Fig. \ref{fig:explanation_ood} left) are called in-distribution (ID), while other environments with distractor objects (Fig. \ref{fig:explanation_ood} middle) or unfamiliar furniture (Fig. \ref{fig:explanation_ood} right) are called out-of-distribution (OOD). 
This domain shift is particularly noticeable with RGB images but might be suppressed with other modalities, such as depth images.
However, the semantics provided by RGB images are crucial for handling task-specific objects among various ones.
Thus, the policy should mainly focus on task-related regions in visual observations and avoid overreacting to domain shifts in non-task-related regions.

In the previous study \cite{cite:see_from_hands}, the policy focuses on task-related viewpoints by uniquely defining from multiple viewpoints to improve robustness to domain shift.
The viewpoints that underwent minimal changes under domain shift were considered important for the task, and a variational information bottleneck (VIB) was added to the back end of the image encoder for the other viewpoints to filter the information.
The study concluded that an in-hand viewpoint is more robust than third-person viewpoints against changes in robot position, distractor objects, and background.
However, the study did not aim to improve robustness to occlusion, and a uniquely defined viewpoint is not necessarily suitable for all occluded situations as mentioned above.
Additionally, just focusing on one unique defined viewpoint is not robust enough to address domain shift.
The policy requires a mechanism that can focus on not only task-related viewpoints but also their spatial regions.

In another previous study \cite{cite:rt1, cite:rosie} used a large dataset to enable learning of task-related regions in the policy for improving robustness to domain shift.
A large dataset is collected through long hours of teleoperated expert demonstration and photo-realistic augmentation using a diffusion model.
By learning the dataset with a policy that includes spatial attention, task-related regions can be focused on.
However, the study was not conducted in a setting with multiple viewpoints, and the performance might decreased for a single internal viewpoint, which can be easily occluded in MMs.
Additionally, although augmentation by the diffusion model is photo-realistic, it has high time and computation demands, which increases in proportion to the number of viewpoints.

In this paper, we propose a robust imitation learning method for MMs that focuses on task-related viewpoints and their spatial regions when observing multiple viewpoints.
First, we propose an attention mechanism for the multiple viewpoints and their spatial regions.
Second, to facilitate the learning of attention mechanism, we propose a method of fast and low computational resource augmentation using fractal texture for non-task-related regions.
This approach allows the policy to focus on task-related regions across multiple viewpoints and improves the robustness of occlusion and domain shift.
Comparison of our results for different tasks and environments with those of previous studies revealed that our proposed method improves the success rate by up to 29.3 points.
We also conduct ablation studies using our proposed method.
Learning task-related viewpoints from the multiple viewpoints dataset increases robustness to occlusion than using a uniquely defined viewpoint.
Focusing on task-related regions contributes to up to a 33.3-point improvement in the success rate against domain shift.

\begin{figure*}[t]
    \begin{center}
        \vspace{2mm}
        \includegraphics[width=0.775\linewidth]{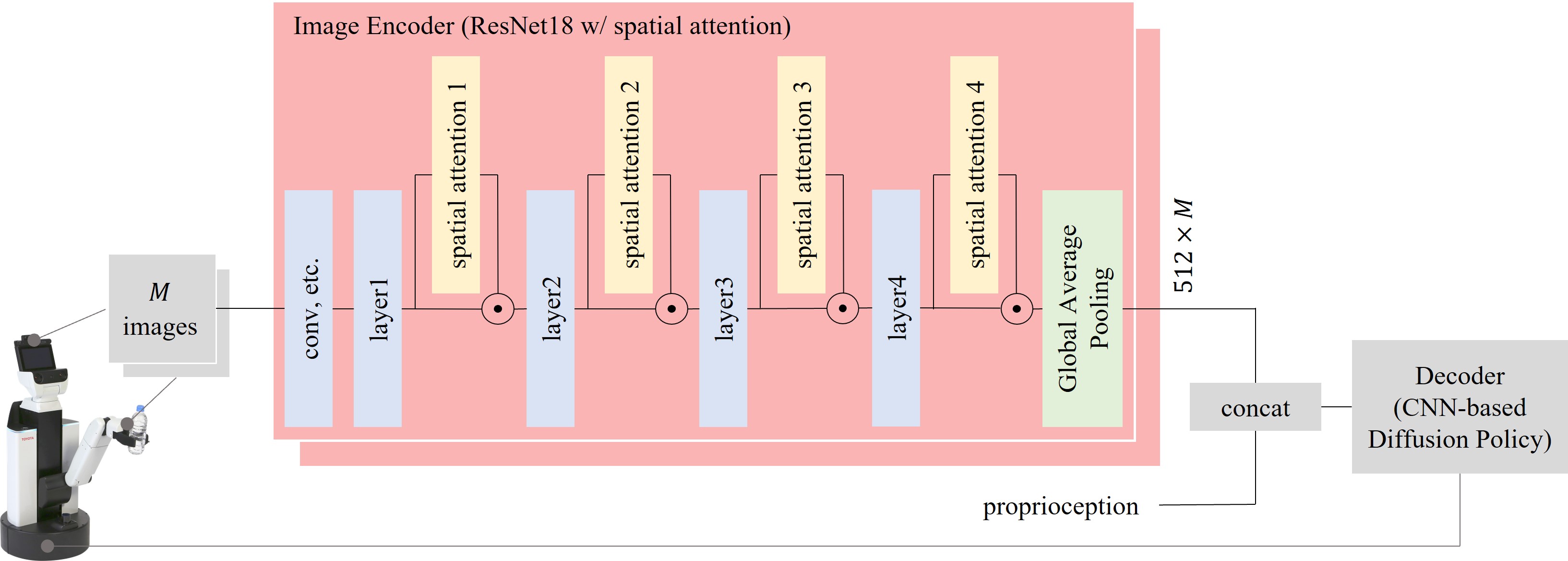}
        \caption{Attention mechanism for multiple viewpoints and their spatial regions. By weighting the features with spatial attention, the information of task-related viewpoints and their spatial regions are extracted in image encoders from multiple visual observations. Since spatial attention is the learnable parameter, our method can learn task-related viewpoints from dataset instead of uniquely defined by hand-craft.}
        \label{fig:proposal_architecture}
    \end{center}
\end{figure*}

\begin{figure*}[t]
    \begin{center}
        \includegraphics[width=0.775\linewidth]{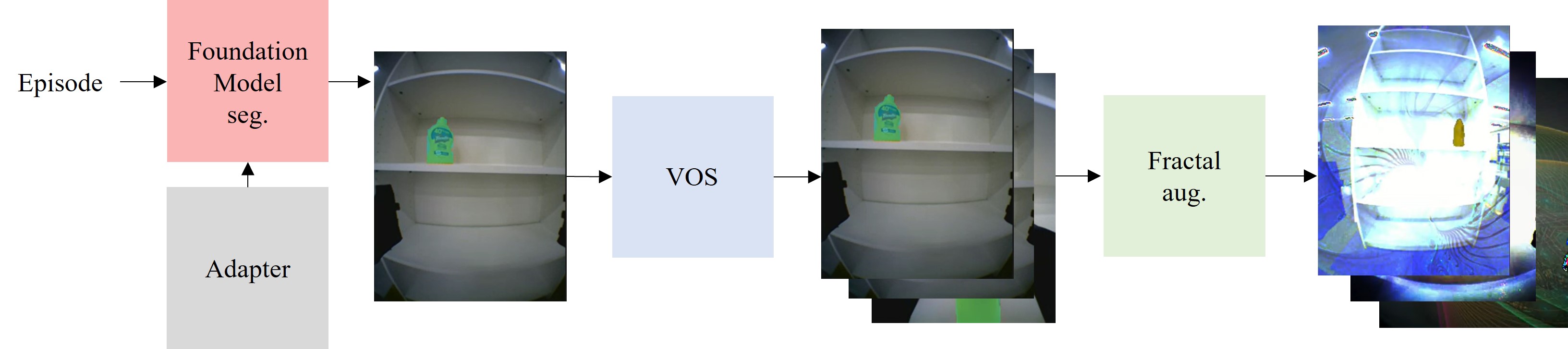}
        \caption{Processing steps of fast and low computational resource augmentation using fractal texture. By detecting and tracking task-related regions, non-task-related regions are augmented with fractal textures. The augmentation facilitates the learning of attention mechanism that focuses strongly on task-related regions which less changed, rather than non-task-related regions that are changed greater with fractal texture.}
        \label{fig:proposal_augmentation}
        \vspace{-2mm}
    \end{center}
\end{figure*}

\section{Related Work}
\label{sec:related_work}

\subsection{Mobile Manipulator}
\label{sec:related_work_mm}

Mobile manipulators have the ability to move and operate in various locations owing to their mobile base.
Moreover, the mobile base allows them to change their posture while they are performing tasks.
Therefore, compared to a manipulator that does not have a mobile base, MMs have the potential to perform a variety of tasks \cite{cite:hsr_toyota_iros, cite:hsr_toyota_robomech}.
The intelligence of MMs has been realized with hand-craft designed algorithms that can be roughly categorized into perception, decision-making, and control algorithms \cite{cite:hsr_ono}.
However, the implementation by skilled engineers requires a considerable amount of time \cite{cite:hsr_luis}.
Additionally, even with the time investment, it is not possible to implement solutions for modeling difficult algorithms by humans such as a manipulator cooking something \cite{cite:aloha}.
Therefore, an MM is expected to learn behaviors through mimicking and trial-and-error by itself using robot learning \cite{cite:bcz, cite:rt1, cite:rt2, cite:mobile_aloha}.

\subsection{Robot Learning}
\label{sec:related_work_rl}

Robots learn behavior through reinforcement learning, which has been studied extensively \cite{cite:mt_opt, cite:visual_foresight}.
In the learning process, the robot is forced to actuate on random or unstable policies to collect data.
This approach carries the risk of malfunction, and its application has been limited to problems where the environment can be automatically reset after each episode.
In recent years, robots have been expected to learn behaviors by imitation learning \cite{cite:see_from_hands, cite:rt1, cite:rosie, cite:aloha, cite:mobile_aloha}.
However, as mentioned in Section \ref{sec:intoroduction}, to the best of the authors’ knowledge, there are no study on robust policies for MMs, which are prone to occlusion and domain shift.

Following the recent success of foundation models in language and image fields, robot learning has also been improved in terms of architecture and data volume.
In terms of architecture, there are many studies on transformer-based policies \cite{cite:rt1, cite:rt2, cite:aloha}.
There are also studies on conditional diffusion models that generate robot trajectories \cite{cite:diffuser, cite:dicision_diffuser, cite:diffusion_policy}.
In terms of data volume, attempts have been made to collect expert demonstrations through long hours of teleoperations \cite{cite:rt1}.
However, it is not practical to collect expert demonstrations of MMs for all situations with teleoperations.
Therefore, there is a need to improve data efficiency by reducing the domain shift which non-task related things by, for example, reducing the dimension of embedding using spatial attention \cite{cite:token_learner}, which is also used in the foundation model.

\section{Preliminaries}
\label{sec:preliminaries}

\textbf{Definitions of Terms:}
We consider the scenario where an MM solves a given task by interacting in the environment.
Specifically, we consider that MMs with parallel grippers manipulate objects as shown in Fig. \ref{fig:explanation_occlusion}.
In this setting, we define \emph{task-related regions} as the spatial region of objects related to the task in the camera image.
The other regions are defined as \emph{non-task-related regions}.
For example, in the pick-red-can task shown in Fig. \ref{fig:explanation_ood}, task-related regions are the pixels corresponding to the red can, and non-task-related regions are any other pixels.

\textbf{Imitation Learning:}
We consider imitation learning in which a robot mimics an expert demonstration and acquires a policy using machine learning.
The expert demonstration is a dataset \( \mathcal{D} = \{ ( {o_{\rm{expert}}}^{(n)}_{t}, {a_{\rm{expert}}}^{(n)}_{t} )^{T^{(n)}}_{t=0} \}^{N}_{n=0} \) of tasks successfully solved by teleoperation.
\(o_{\rm{expert}}\) and \(a_{\rm{expert}}\) are the robot's observations and actions acquired during teleoperation, \(T\) is the timestep of each episode, and \(N\) is the number of episodes.
The policy \(\pi(a|o)\) is trained by supervised learning using dataset \(\mathcal{D}\) to minimize the error between predicted action \(a_{\rm{pred}}\) and \(a_{\rm{expert}}\) when \(o_{\rm{expert}}\) is given as the input.
At test time, the robot works by continuing the process of inputting the current observation \(o\) and predicting action \(a\) using policy \(\pi(a|o)\).

\textbf{Observations and Actions:}
We focus on MMs with internal multiple sensors as shown in Fig. \ref{fig:explanation_occlusion}.
As a type of observation, we consider in-hand viewpoint \(o_{h}\), first-person viewpoint \(o_{f}\) and proprioception \(o_{p}\).
The action \(a\) is the relative movement of the end-effector frame.
The low-level controller receives \(a\) from the policy and solves inverse kinematics to convert it into a joint position command value.


\begin{table*}[tb]
    \begin{center}
        \vspace{2mm}
        \caption{Hyperparameters for DP \cite{cite:diffusion_policy}.}
        \begin{tabular}{rrrrrrrrrrrr}
            \toprule
            Ctrl    & To  & Ta    & Tp    & ImgRes    & CropRes   & \#D-Params    & \#V-Params    & Lr    & WDecay    & D-Inters Train    & D-Inters Eval   \\
            \midrule
            Pos     & 2   & 8     & 16    & 2x320x240 & 2x288x216 & 67 millions   & 22 millions   & 1e-4  & 1e-6      & 100               & 8               \\
            \bottomrule
        \end{tabular}
        \label{tab:dp_hyperparameter}
        \vspace{-2mm}
    \end{center}
\end{table*}

\section{Proposed Method}
\label{sec:proposal_method}

We propose a robust imitation learning method for MMs that focuses on task-related viewpoints and their spatial regions when observing multiple viewpoints.
First, we propose an attention mechanism for multiple viewpoints and their spatial regions.
Second, to facilitate the learning of spatial attention, we propose a method of fast and low computational resource augmentation using fractal texture for non-task-related regions.
These methods enable the policy to focus on task-related regions across multiple viewpoints and improve the robustness of occlusion and domain shift.

\subsection{Attention Mechanism for Multiple Viewpoints and Their Spatial Regions for Imitation Learning Policy}
\label{sec:proposal_method_sa}

Our proposed method modifies a common architecture of imitation learning with visual observation, which combines a network of encoders and decoders.
Here we illustrate our proposed method using an example of an encoder and decoder policy, Diffusion Policy (DP) \cite{cite:diffusion_policy} (Fig. \ref{fig:proposal_architecture}).
As a first modification, our proposed method focuses on task-related regions to improve the robustness of domain shift.
Following a previous study \cite{cite:rt1} that conducted a single viewpoint, our proposed method incorporates spatial attention inside the image encoder as shown in Fig. \ref{fig:proposal_architecture}.
The spatial attention predicts weights \(w(b, h, w, 1)\) using multiple convolution layers for a feature \(f(b, h, w, c)\) on a convolutional neural network, expands it to channel dimension, and finally takes the inner product of \(f(b, h, w, c)\) and \(w(b, h, w, c)\).
The features are weighted to task-related regions.
As a second modification, our proposed method handles multiple viewpoints to improve the robustness against occlusion.
Our proposed method includes per-viewpoint image encoders instead of a shared encoder to focus spatial regions in each viewpoint as shown in Fig. \ref{fig:proposal_architecture}.
The image encoders for \(M\) viewpoints have parameters corresponding to each viewpoint.

Our proposed method focuses on the optimal viewpoints of a task by keeping only the information of task-related viewpoints from multiple viewpoints using the attention mechanism.
Contrary to a previous study \cite{cite:see_from_hands} that uniquely defined optimal single viewpoints, our method can learn task-related viewpoints from a dataset. Thus, it can be applied for tasks that are performed under various conditions.
Additionally, our proposed method acquires robust visual embedding against domain shift by keeping only the information of task-relevant spatial regions using the attention mechanism.
The embedding is robust against large distribution shifts in visual observation caused by distractor objects or unknown furniture as shown in Fig. \ref{fig:explanation_ood}.

\subsection{Fast and Low Computational Resource Augmentation using Fractal Texture}
\label{sec:proposal_method_aug}

To facilitate the learning of attention mechanism, we propose a method of augmenting non-task related regions with fractal texture (Fig. \ref{fig:proposal_augmentation}).
The attention mechanism focuses strongly on task-related regions that have undergone minimal changes rather than non-task-related regions that have undergone considerable changes after application of fractal texture.
Contrary to the previous study \cite{cite:rosie}, our proposed method has minimal time and computational resources, making it suitable for augmenting multiple viewpoints with a larger volume of images compared to a single viewpoint.

First, our proposed method detects task-related regions in visual observations \(o_{h}\) and \(o_{f}\) at \(t=0\) in dataset \(\mathcal{D}\) as shown Fig. \ref{fig:proposal_augmentation} left.
Foundation models adapted to robotic training data, such as a combination of FastSAM \cite{cite:fastsam} and CLIP-Adapter \cite{cite:clip_adapter}, use text prompts to detect task-specific objects.
(Note that FastSAM is adapted to in-distribution training data and requires hyperparameters related to segmentation granularity to be tuned according to the input, which make it less effective in detecting with out-of-distribution data during execution.)
Second, the task-related regions are tracked from \(t=0\) to \(t=T(n)\) as shown Fig. \ref{fig:proposal_augmentation} middle.
XMem \cite{cite:XMem} is the video object segmentation (VOS) method used for tracking.
These steps generate non-task-related region masks, which is complement of task-related regions, for all times during all the episodes.
Finally, augmentation is applied to non-task related regions using PixMix \cite{cite:pixmix} as shown Fig. \ref{fig:proposal_augmentation} right.
Objects with patterns ranging from simple contours, such as a box, to complex contours, such as fern leaves, appear in non-task-related regions.
Fractals can be controlled by the complexity of their patterns with parameters and have a structure in which a part is self-similar to the whole, for example, a pattern similar to the original appears even after repeated enlargements.
Therefore, the augmented dataset simulates artificial and natural objects by various patterns of fractals and is as effective as a large dataset with various backgrounds collected through long hours of teleoperation.

\section{Experiments}
\label{sec:experiments}

\begin{figure}[t]
    \begin{center}
        \includegraphics[width=0.95\linewidth]{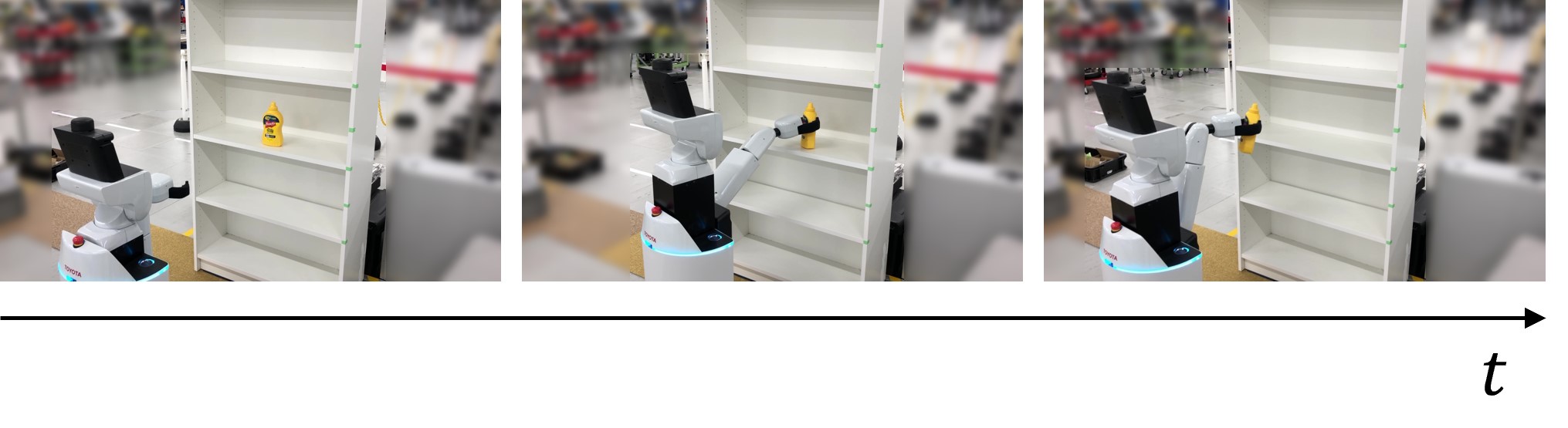}
        \caption{Overview of the pick-bottle-from-shelf task. Figures are lined in time-step order from left to right. Left: the MM started with \( o_{h} \) and \( o_{f} \) facing the bottle placed on the shelf. Middle: the MM moved the mobile base and arm to reach the bottle. Right: the MM picked up the bottle from the shelf.}
        \label{fig:pick_bottle_from_shelf_sequence}
        \vspace{-2mm}
    \end{center}
\end{figure}

\begin{figure}[t]
    \begin{center}
        \includegraphics[width=0.95\linewidth]{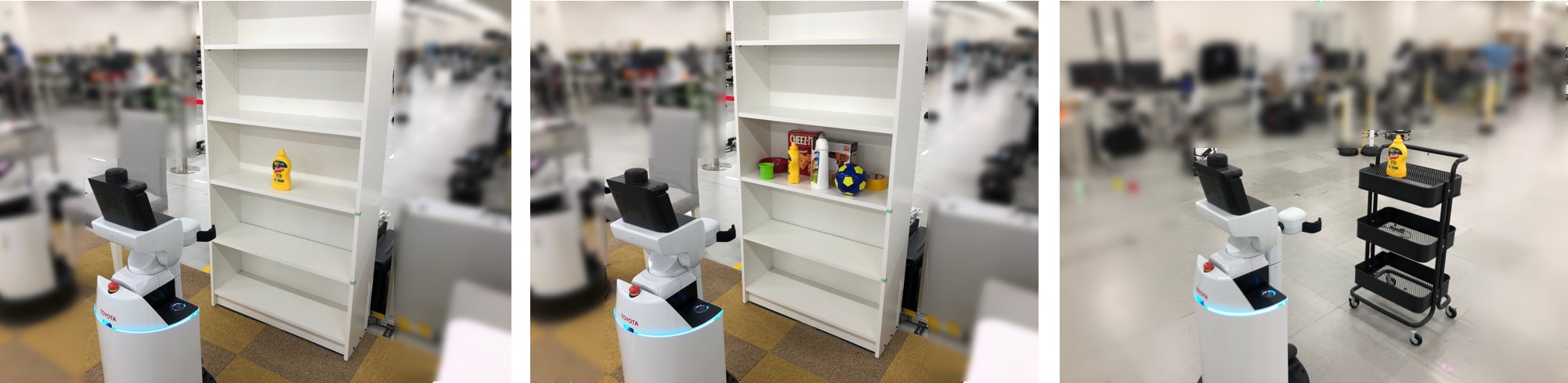}
        \caption{List of pick-bottle-from-shelf task environments. Left: In distribution (ID). Middle: Out of distribution due to the distractor objects (OOD-distractor). Right: Out of distribution due to the unknown shelf (OOD-shelf).}
        \label{fig:pick_bottle_from_shelf_environments}
        \vspace{-2mm}
    \end{center}
\end{figure}



To validate the effectiveness of our proposed method, we performed experiments by varying the task and environment.
In the following section, we describe the experimental settings, compare our results with those of previous studies, and describe ablation studies to answer the following three questions.

\begin{enumerate}
    \item What is the effect of the multiple viewpoints?
    \item How can MM focus on task-related viewpoints?
    \item How can MM focus on task-related regions?
\end{enumerate}

\subsection{Settings: Tasks and Environments}
\label{sec:experiments_task_env}

\textbf{pick-bottle-from-shelf:}
Figure \ref{fig:pick_bottle_from_shelf_sequence} shows the sequence of the pick-bottle-from-shelf task in time-step order.
The MM started by observing the bottle placed on the shelf with \(o_h\) or \(o_f\) (Fig. \ref{fig:pick_bottle_from_shelf_sequence} left), reached the bottle (Fig. \ref{fig:pick_bottle_from_shelf_sequence} middle), and picked up the bottle from the shelf (Fig. \ref{fig:pick_bottle_from_shelf_sequence} right).
If the MM could lift the bottle off the shelf board and carry it out of the shelf, it was successful in completing the task.
In this task, \(o_h\) had no occlusion to observe the bottle and \(o_f\) had occlusion due to the body of the MM (Fig. \ref{fig:pick_bottle_from_shelf_viewpoint} right). 
Thus, \(o_h\) had more information on the visuomotor policy.

Three environmental variations are shown in Fig. \ref{fig:pick_bottle_from_shelf_environments}.
The ID environment in Fig. \ref{fig:pick_bottle_from_shelf_environments} left was the same as when the expert demonstrations were collected.
The OOD-distractor environment in Fig. \ref{fig:pick_bottle_from_shelf_environments} middle had the same shelf as when the expert demonstration was collected, but there were distractor objects around the bottle.
The OOD-shelf environment on Fig. \ref{fig:pick_bottle_from_shelf_environments} right had an unknown shelf.
The expert demonstrations were collected 50 times in the ID environment, and each environment was tested 15 times while varying the positions of the MM, the bottle, and distractor objects.

\begin{figure}[t]
    \begin{center}
        \vspace{2mm}
        \includegraphics[width=0.95\linewidth]{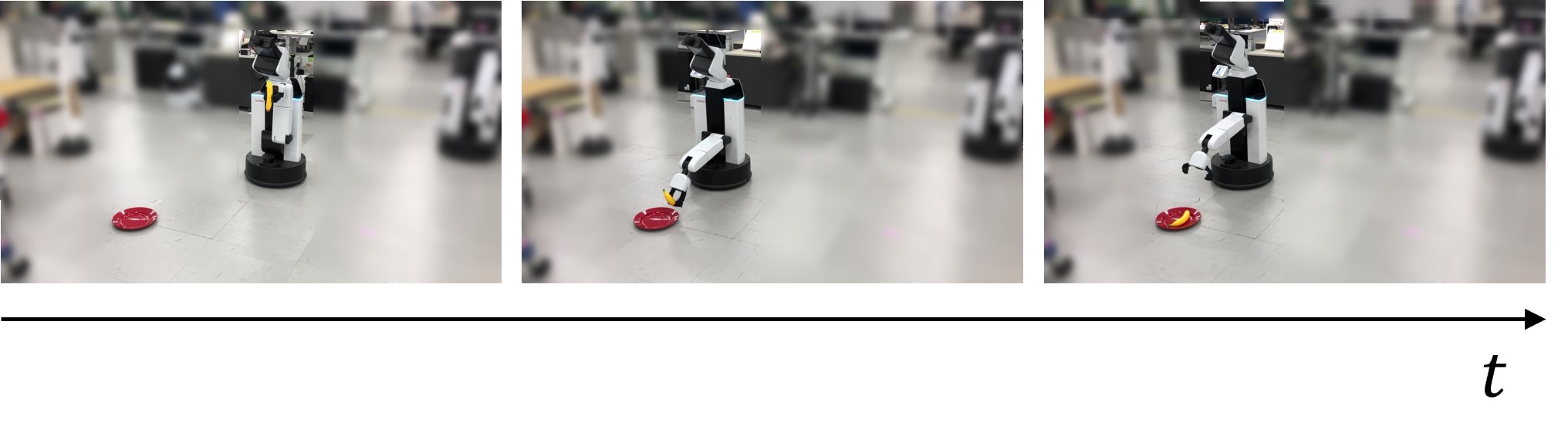}
        \caption{Overview of the place-banana-on-plate task. Figures are lined in time-step order from left to right. Left: the MM started by grasping the banana with \( o_{f} \) facing the plate on the floor. Middle: the MM moved the mobile base and arm to reach the plate. Right: the MM placed and left the banana on the plate.}
        \label{fig:place_banana_on_plate_sequence}
        \vspace{-2mm}
    \end{center}
\end{figure}

\textbf{place-banana-on-plate:}
Figure \ref{fig:place_banana_on_plate_sequence} shows the sequence of the place-banana-on-plate task in time-step order.
The MM started by observing the plate placed on the floor with \(o_f\) (Fig. \ref{fig:place_banana_on_plate_sequence} left), reached the plate (Fig. \ref{fig:place_banana_on_plate_sequence} middle), and placed the banana on the plate (Fig. \ref{fig:place_banana_on_plate_sequence} right).
If the MM could place the banana on the plate without touching the floor, it was successful in completing the task.
At the beginning of the time-step, \(o_f\) had no occlusion to observe the plate, while \(o_h\) only observed the banana and the background (Fig. \ref{fig:place_banana_on_plate_viewpoint} left).
However, in the late stage of the time-step, \(o_f\) experienced occlusion due to the body of the MM, and \(o_h\) could observe the plate (Fig. \ref{fig:place_banana_on_plate_viewpoint} right).
Thus, both \(o_f\) and \(o_h\) had the information on the visuomotor policy such as the position of the plate.

\begin{figure}[t]
    \begin{center}
        \includegraphics[width=0.825\linewidth]{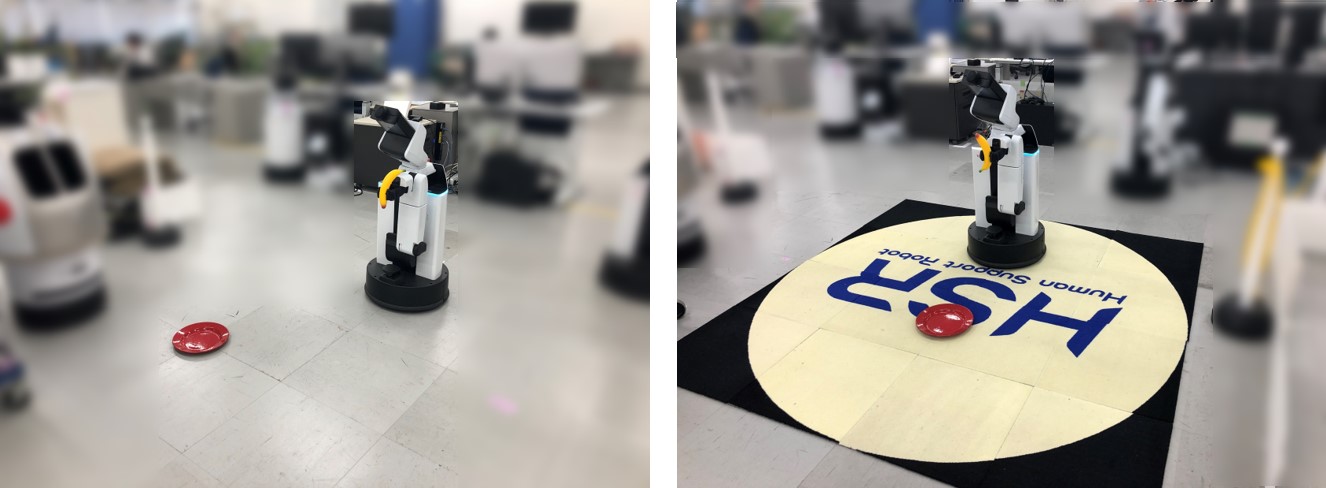}
        \caption{List of place-banana-on-plate task environments. Left: In distribution (ID). Right: Out of distribution due to the unknown floor texture (OOD-shelf).}
        \label{fig:place_banana_on_plate_environments}
        \vspace{-2mm}
    \end{center}
\end{figure}

Two environmental variations are shown in Fig \ref{fig:place_banana_on_plate_environments}.
The ID environment in Fig. \ref{fig:place_banana_on_plate_environments} left was the same as when the expert demonstrations were collected.
The OOD-floor environment on Fig. \ref{fig:place_banana_on_plate_environments} right had an unknown floor texture.
The expert demonstrations were collected 50 times in the ID environment, and each environment was tested 15 times while varying the positions of the MM and the plate.

\subsection{Settings: Implementations}
\label{sec:experiments_impl}

The following three methods were implemented for comparison: (1) Previous study: Original DP \cite{cite:diffusion_policy}; (2) previous study: DP with VIB on each back end of image encoders and augmentation (Aug) \cite{cite:see_from_hands}, and (3) proposed method: DP with attention mechanism (AM) and augmentation.
The DP hyperparameters are shown in Table \ref{tab:dp_hyperparameter}.
The architecture of VIB and spatial attention are shown in Table \ref{tab:vib_description} and Table \ref{tab:sa_description} respectively.
\(\beta\) used in the VIB loss function was set to 1.0.
We used the Toyota Human Support Robot (HSR) \cite{cite:hsr_toyota_robomech} as the MM.

\begin{table}[t]
    \begin{center}
        \vspace{2mm}
        \caption{Architecture of VIB}
        \begin{tabular}{llc}
            \toprule
                    & \multicolumn{1}{c}{Layer Description} & \multicolumn{1}{c}{Output Tensor Dim.}    \\
            \midrule
            \#0     & Input feature                         & 512                                       \\
            \addlinespace[1mm]
            \multicolumn{3}{c}{Encoder}                                                                 \\
            \addlinespace[1mm]
            \#1     & Linear + ReLU                         & 256                                       \\
            \#2     & Linear + ReLU                         & 256                                       \\
            \addlinespace[1mm]
            \multicolumn{3}{c}{\(\mu\) layer}                                                           \\
            \addlinespace[1mm]
            \#3-1   & Linear                                & 128                                       \\
            \addlinespace[1mm]
            \multicolumn{3}{c}{\(\sigma\) layer}                                                        \\
            \addlinespace[1mm]
            \#3-2   & Linear + Softplus                     & 128                                       \\
            \addlinespace[1mm]
            \multicolumn{3}{c}{Latent layer}                                                            \\
            \addlinespace[1mm]
            \#4     & \(\mu + \sqrt{\sigma} \times randn\)  & 128                                       \\
            \addlinespace[1mm]
            \multicolumn{3}{c}{Decoder}                                                                 \\
            \addlinespace[1mm]
            \#5     & Linear                                & 512                                       \\
            \bottomrule
        \end{tabular}
        \label{tab:vib_description}
    \end{center}
\end{table}

\begin{table}[t]
    \begin{center}
        \caption{Architecture of spatial attention}
        \begin{tabular}{llc}
            \toprule
                & \multicolumn{1}{c}{Layer Description} & \multicolumn{1}{c}{Output Tensor Dim.}    \\
            \midrule
            \#0 & Input feature                         & \(c \times h \times w\)                   \\
            \#1 & Conv2d + ReLU                         & \(c \times h \times w\)                   \\
            \#2 & Conv2d + Sigmoid                      & \(1 \times h \times w\)                   \\
            \#3 & expand (Channel Dim.)                 & \(c \times h \times w\)                   \\
            \bottomrule
        \end{tabular}
        \label{tab:sa_description}
    \end{center}
\end{table}

\begin{table}[t]
    \begin{center}
        \vspace{2mm}
        \caption{Overall results of the pick-bottle-from-shelf task}
        \begin{tabular}{lrrr}
            \toprule
            \multirow{3}{*}{method}                                         & \multicolumn{3}{c}{Success Rate [\%]}                                                                                         \\
                                                                            & \multicolumn{1}{c}{ID}    & \begin{tabular}{c}OOD\\(distracter)\end{tabular}  & \begin{tabular}{c}OOD\\(shelf)\end{tabular}   \\
            \midrule
            Original \cite{cite:diffusion_policy}                           & 86.7                      & 46.7                                              & 0.0                                           \\
            VIB(\(z_h\)) + Aug \cite{cite:see_from_hands}                   & 0.0                       & 6.7                                               & 6.7                                           \\
            VIB(\(z_f\)) + Aug \cite{cite:see_from_hands}                   & \textbf{93.3}             & 73.3                                              & 86.7                                          \\
            VIB(\(z_h\)) + VIB(\(z_f \)) + Aug \cite{cite:see_from_hands}   & 86.7                      & 40.0                                              & 60.0                                          \\
            \textbf{Proposal: AM + Aug}                                     & \textbf{93.3}             & \textbf{80.0}                                     & \textbf{100.0}                                \\
            \bottomrule
        \end{tabular}
        \label{tab:overall_pick}
    \end{center}
\end{table}

\begin{table}[tb]
    \begin{center}
        \caption{Overall results of the place-banana-on-plate task}
        \begin{tabular}{lrr}
            \toprule
            \multirow{3}{*}{method}                                         & \multicolumn{2}{c}{Success Rate [\%]}                                     \\
                                                                            & \multicolumn{1}{c}{ID}    & \begin{tabular}{c}OOD\\(floor)\end{tabular}   \\
            \midrule
            Original \cite{cite:diffusion_policy}                           & \textbf{100.0}            & 73.3                                          \\
            VIB(\(z_h\)) + Aug \cite{cite:see_from_hands}                   & 60.0                      & 33.3                                          \\
            VIB(\(z_f\)) + Aug \cite{cite:see_from_hands}                   & 66.7                      & 20.0                                          \\
            VIB(\(z_h\)) + VIB(\(z_f \)) + Aug \cite{cite:see_from_hands}   & 13.3                      & 0.0                                           \\
            \textbf{Proposal: AM + Aug}                                     & 93.3                      & \textbf{86.7}                                 \\
            \bottomrule
        \end{tabular}
        \label{tab:overall_place}
    \end{center}
\end{table}

\begin{table}[b]
    \begin{center}
        \caption{Ablation study: effect of multiple viewpoints\\in the pick-bottle-from-shelf task}
        \begin{tabular}{lrrr}
            \toprule
            \multirow{3}{*}{observation}            & \multicolumn{3}{c}{Success Rate [\%]}                                                                                         \\
                                                    & \multicolumn{1}{c}{ID}    & \begin{tabular}{c}OOD\\(distracter)\end{tabular}  & \begin{tabular}{c}OOD\\(shelf)\end{tabular}   \\
            \midrule
            \(o_h, o_p\)                            & \textbf{100.0}            & \textbf{80.0}                                     & 80.0                                          \\
            \(o_f, o_p\)                            & 20.0                      & 6.7                                               & 6.7                                           \\
            \textbf{Proposal: \(o_h, o_f, o_p\)}    & 93.3                      & \textbf{80.0}                                     & \textbf{100.0}                                \\
            \bottomrule
        \end{tabular}
        \label{tab:viewpoint_pick}
    \end{center}
\end{table}

\begin{table}[b]
    \begin{center}
        \caption{Ablation study: effect of multiple viewpoints\\in the place-banana-on-plate task}
        \begin{tabular}{lrr}
            \toprule
            \multirow{3}{*}{observation}            & \multicolumn{2}{c}{Success Rate [\%]}                                     \\
                                                    & \multicolumn{1}{c}{ID}    & \begin{tabular}{c}OOD\\(floor)\end{tabular}   \\
            \midrule
            \(o_h, o_p\)                            & 26.7                      & 26.7                                          \\
            \(o_f, o_p\)                            & 86.7                      & 66.7                                          \\
            \textbf{Proposal: \(o_h, o_f, o_p\)}    & \textbf{93.3}             & \textbf{86.7}                                 \\
            \bottomrule
        \end{tabular}
        \label{tab:viewpoint_place}
    \end{center}
\end{table}

\subsection{Overall Results}
\label{sec:experiments_overall}

Comparison of the results of previous studies and our proposed method in the pick-bottle-from-shelf task and place-banana-on-plate tasks are shown in Table \ref{tab:overall_pick} and Table \ref{tab:overall_place} respectively.
As described in Section \ref{sec:experiments_impl}, the results of two previous studies were compared with the results of our proposed method.
All methods had three observations \(o_h, o_f, o_p\).
Following the previous studies, VIB was applied for either or both of encoded \(o_h\) (\(z_h\)) and \(o_f\) (\(z_f\)).

Compared to the original DP, our proposed method improved the success rate by 6.6 points in the pick-bottle-from-shelf ID environment (Table \ref{tab:overall_pick}).
No improvement was observed in the place-banana-on-plate ID environment (Table \ref{tab:overall_place}).
A large success rate difference appeared in OOD environments. Our proposed method demonstrated improvements of 33.3 points in pick-bottle-from-shelf OOD-distractor, 100.0 points in pick-bottle-from-shelf OOD-shelf, and 13.4 points in place-banana-on-plate OOD-floor respectively (Table \ref{tab:overall_pick} and \ref{tab:overall_place}).
Even after including degradation, which was limited to one failure in 15 tests of one environment, our proposed method improved the average success rate for the five environments by 29.3 points.
The improvement in the 4/5 environment was due to the attention mechanism and augmentation being robust to domain shift of non-task related regions, as described in Section \ref{sec:proposal_method}.

Compared to the original DP, VIB exhibited the same trend reported in a previous study and improved the success rate in the pick-bottle-from-shelf task in the order of VIB(\(z_f\)), VIB(\(z_h, z_f\)) and VIB(\(z_h\)) (Table \ref{tab:overall_pick}).
However, VIB exhibited a different trend, and no improvement was observed in the place-banana-on-plate task (Table \ref{tab:overall_place}).
The reason for the different trend was that VIB uniquely defined a viewpoint as mentioned in Section \ref{sec:intoroduction} and could not handle cases of changing task-related viewpoints during execution as mentioned in Section \ref{sec:experiments_task_env}.

Compared to VIB(\(z_f\)), which was considered the best in previous studies, our proposed method achieved an equal or higher success rate in all tasks and environments.
Our proposed method demonstrated improvements of 6.7 points in pick-bottle-from-shelf OOD-distractor, 13.3 points in pick-bottle-from-shelf OOD-shelf, 26.6 points in place-banana-on-plate ID, and 66.7 points in place-banana-on-plate OOD-floor respectively (Table \ref{tab:overall_pick} and \ref{tab:overall_place}).
Our proposed method improved the average success rate of the five environments by 22.7 points.
These results were achieved because our proposed method can learn task-related viewpoints from datasets in various situations and could focus on task-related regions as mentioned in Section \ref{sec:proposal_method} (Detailed in Section \ref{sec:experiments_viewpoint} and \ref{sec:experiments_region}).

\begin{figure}[t]
    \begin{center}
        \vspace{2mm}
        \includegraphics[width=0.775\linewidth]{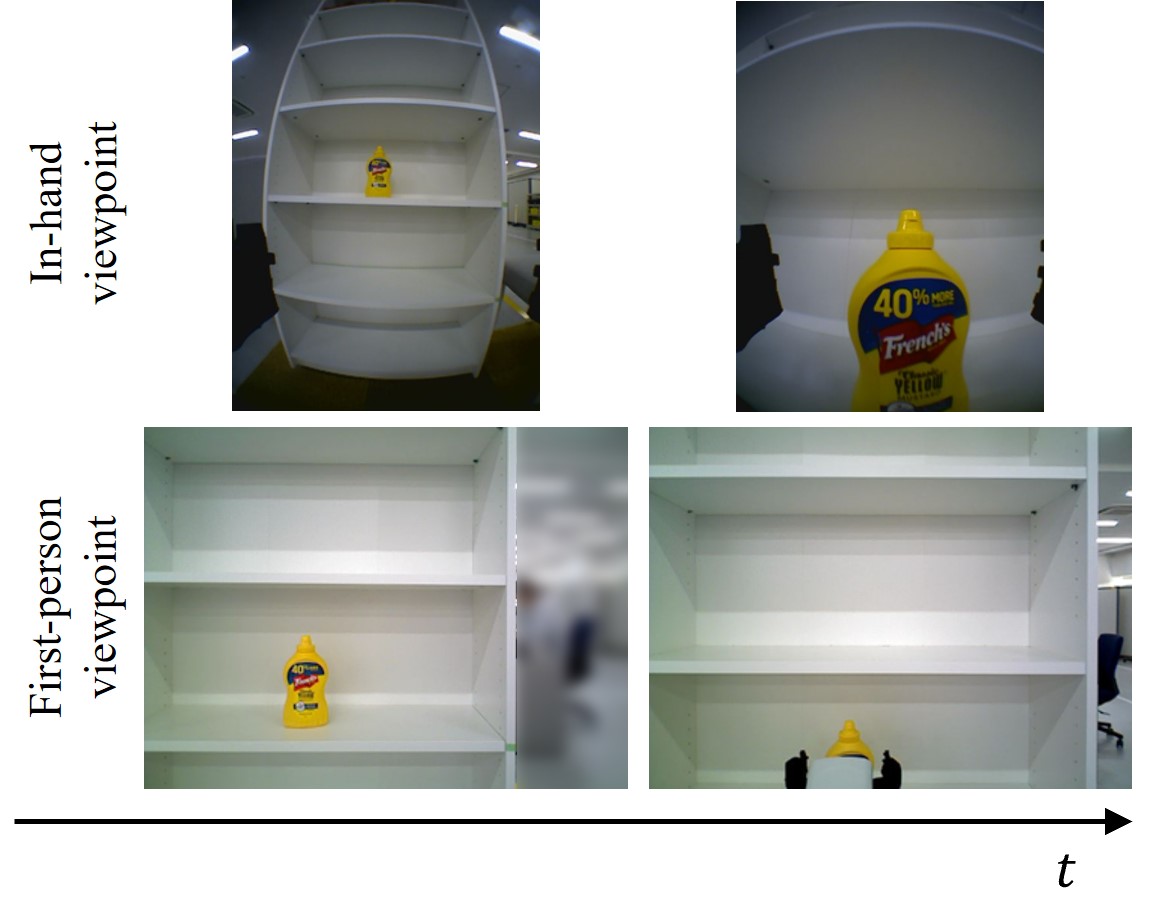}
        \caption{Visual observations in the pick-bottle-from-shelf task. Figures are lined in time-step order from left to right.}
        \label{fig:pick_bottle_from_shelf_viewpoint}
        \vspace{-2mm}
    \end{center}
\end{figure}

\begin{figure}[t]
    \begin{center}
        \includegraphics[width=0.775\linewidth]{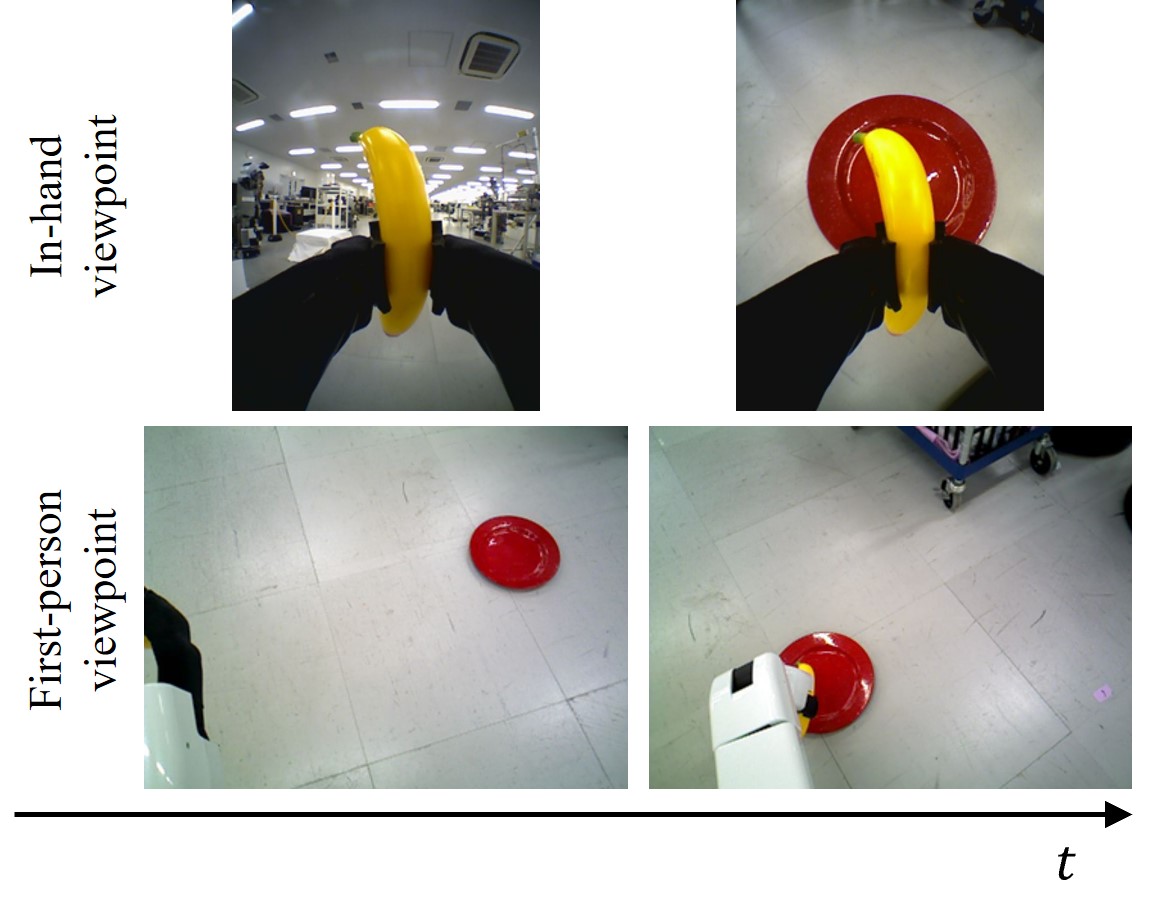}
        \caption{Visual observations in the place-banana-on-plate task. Figures are lined in time-step order from left to right.}
        \label{fig:place_banana_on_plate_viewpoint}
        \vspace{-2mm}
    \end{center}
\end{figure}

\subsection{What is the effect of the multiple viewpoints?}
\label{sec:experiments_multi}

To clarify the effect of multiple viewpoints, the ablation study of viewpoints in the pick-bottle-from-shelf and place-banana-on-plate tasks are shown in Table \ref{tab:viewpoint_pick} and Table \ref{tab:viewpoint_place} respectively.
All methods were based on our proposed method described in section \ref{sec:experiments_impl}, and three types of observations are compared: only in-hand viewpoint \(o_h, o_p\), only first-person viewpoint \(o_f, o_p\), and multiple viewpoints \(o_h, o_f, o_p\).

In the pick-bottle-from-shelf task, the in-hand viewpoint achieved a higher success rate than the first-person viewpoint (Table \ref{tab:viewpoint_pick}).
On the other hand, in the place-banana-on-plate task, the in-hand viewpoint achieved a lower success rate than the first-person viewpoint (Table \ref{tab:viewpoint_place}).
As described in Section \ref{sec:experiments_task_env}, each task had different occluded viewpoints, and use of a single viewpoint did not provide robustness against occlusion (Fig. \ref{fig:pick_bottle_from_shelf_viewpoint} and \ref{fig:place_banana_on_plate_viewpoint}).
Table \ref{tab:viewpoint_distance} shows the stratified analyses of the initial distance between the MM and the plate in the place-banana-on-plate task.
The difference between the in-hand viewpoint and first-person viewpoint was 44.5 points for short distances and 52.4 points for long distances.
For short-distance cases, the MM might observe the plate when the in-hand viewpoint is pointed at the floor in a guesswork manner.
On the contrary, the MM cannot observe the plate for long-distance cases.
Thus, the number of failures increase.

For both the pick-bottle-from-shelf and place-banana-on-plate tasks, the use of multiple viewpoints resulted in higher success rates than the use of in-hand viewpoint or first-person viewpoint (Table \ref{tab:viewpoint_pick} and \ref{tab:viewpoint_place}).
Thus, multiple viewpoints provide robust visual observation even in cases where occlusion occurs, such as when the internal viewpoints of the MM are used.

\begin{table}[t]
    \begin{center}
        \vspace{2mm}
        \caption{Stratified analyses of the initial distance between\\the MM and the plate in the place-banana-on-plate task}
        \begin{tabular}{lrr}
            \toprule
            \multirow{2}{*}{observation}            & \multicolumn{2}{c}{Success Rate [\%]}                                     \\
                                                    & \multicolumn{1}{c}{\(<\) 0.7 [m]} & \multicolumn{1}{c}{\(\geq\) 0.7 [m]}  \\
            \midrule
            \(o_h, o_p\)                            & 44.4                              & 19.0                                  \\
            \(o_f, o_p\)                            & 88.9                              & 71.4                                  \\
            \textbf{Proposal: \(o_h, o_f, o_p\)}    & \textbf{100.0}                    & \textbf{85.7}                         \\
            \bottomrule
        \end{tabular}
        \label{tab:viewpoint_distance}
    \end{center}
\end{table}

\begin{figure}[t]
    \begin{center}
        \includegraphics[width=0.775\linewidth]{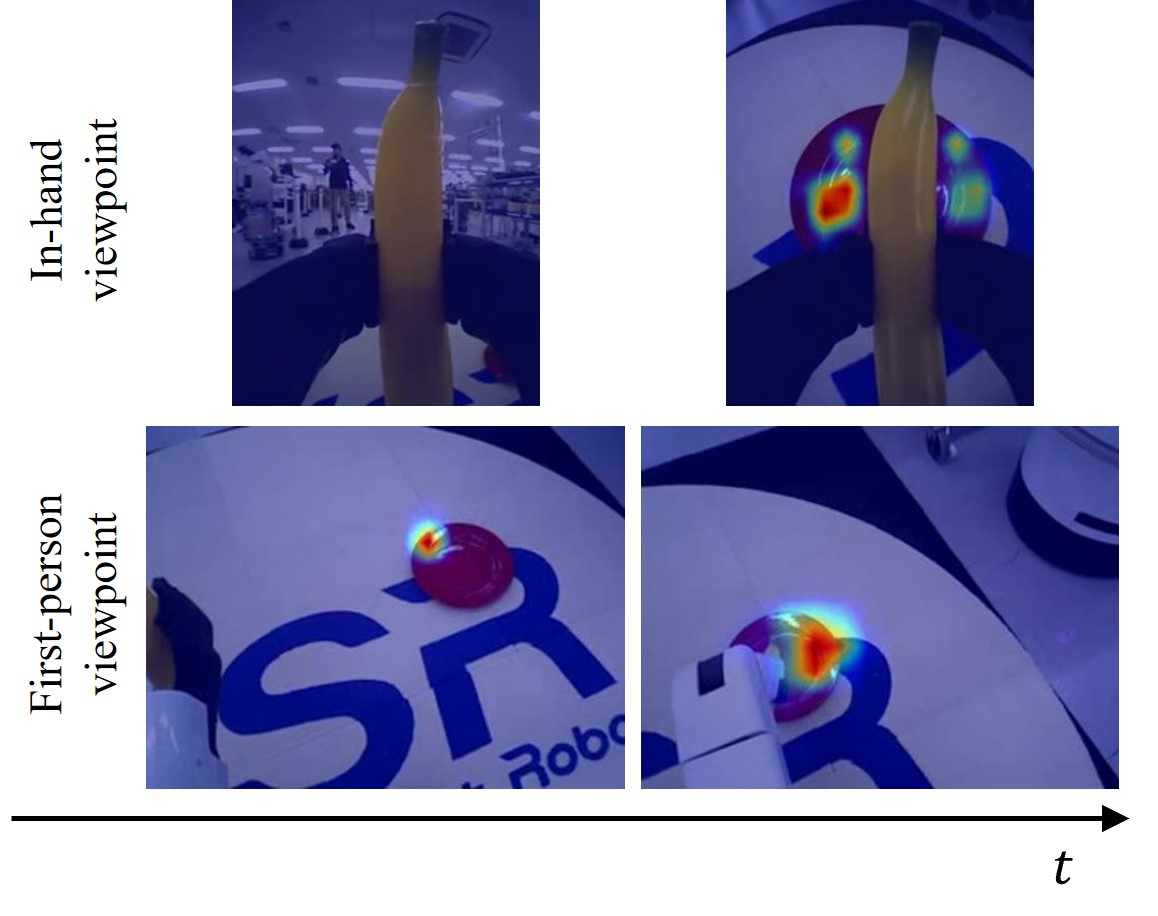} 
        \caption{Visualize pixel attribution on the prediction of our proposed method in the place-banana-on-plate task. Figures are lined in time-step order from left to right. The method was able to focus on the plate, which changes visibility moment by moment.}
        \label{fig:attention_place_banana_on_plate}
        \vspace{-2mm}
    \end{center}
\end{figure}

\subsection{How can MM focus on task-related viewpoint?}
\label{sec:experiments_viewpoint}

As mentioned in Section \ref{sec:experiments_overall}, compared to VIB(\(z_f\)), which was considered the best in previous studies, our proposed method achieved an equal or high success rate in all tasks and environments.
Figure \ref{fig:attention_place_banana_on_plate} shows qualitative analysis with visualize the pixel attribution on the prediction of our proposed method using GradCAM \cite{cite:gradcam} in the place-banana-on-plate task where a difference of success rate was particularly large.
At the beginning of the time-step, the method focused on the plate in \(o_f\), as shown in Fig. \ref{fig:attention_place_banana_on_plate} left.
In the late stage of the time-step, the method focused on the plate in both \(o_h\) and \(o_f\), as shown in Fig. \ref{fig:attention_place_banana_on_plate} right.
Compared to VIB which uniquely defined viewpoints, our proposed method could focus on optimal viewpoints based on the situation and adapt well to various tasks.

\begin{table}[t]
    \begin{center}
        \vspace{2mm}
        \caption{Ablation study of how to focus task-related regions}
        \begin{tabular}{lrrr}
            \toprule
            \multirow{3}{*}{method}     & \multicolumn{3}{c}{Success Rate [\%]}                                                                                         \\
                                        & \multicolumn{1}{c}{ID}    & \begin{tabular}{c}OOD\\(distracter)\end{tabular}  & \begin{tabular}{c}OOD\\(shelf)\end{tabular}   \\
            \midrule
            AM                          & \textbf{100.0}            & 73.3                                              & 0.0                                           \\
            Aug                         & 86.7                      & 66.7                                              & 66.7                                          \\
            \textbf{Proposal: AM + Aug} & 93.3                      & \textbf{80.0}                                     & \textbf{100.0}                                \\
            \bottomrule
        \end{tabular}
        \label{tab:region}
    \end{center}
\end{table}

\begin{figure}[t]
    \begin{center}
        \includegraphics[width=0.6\linewidth]{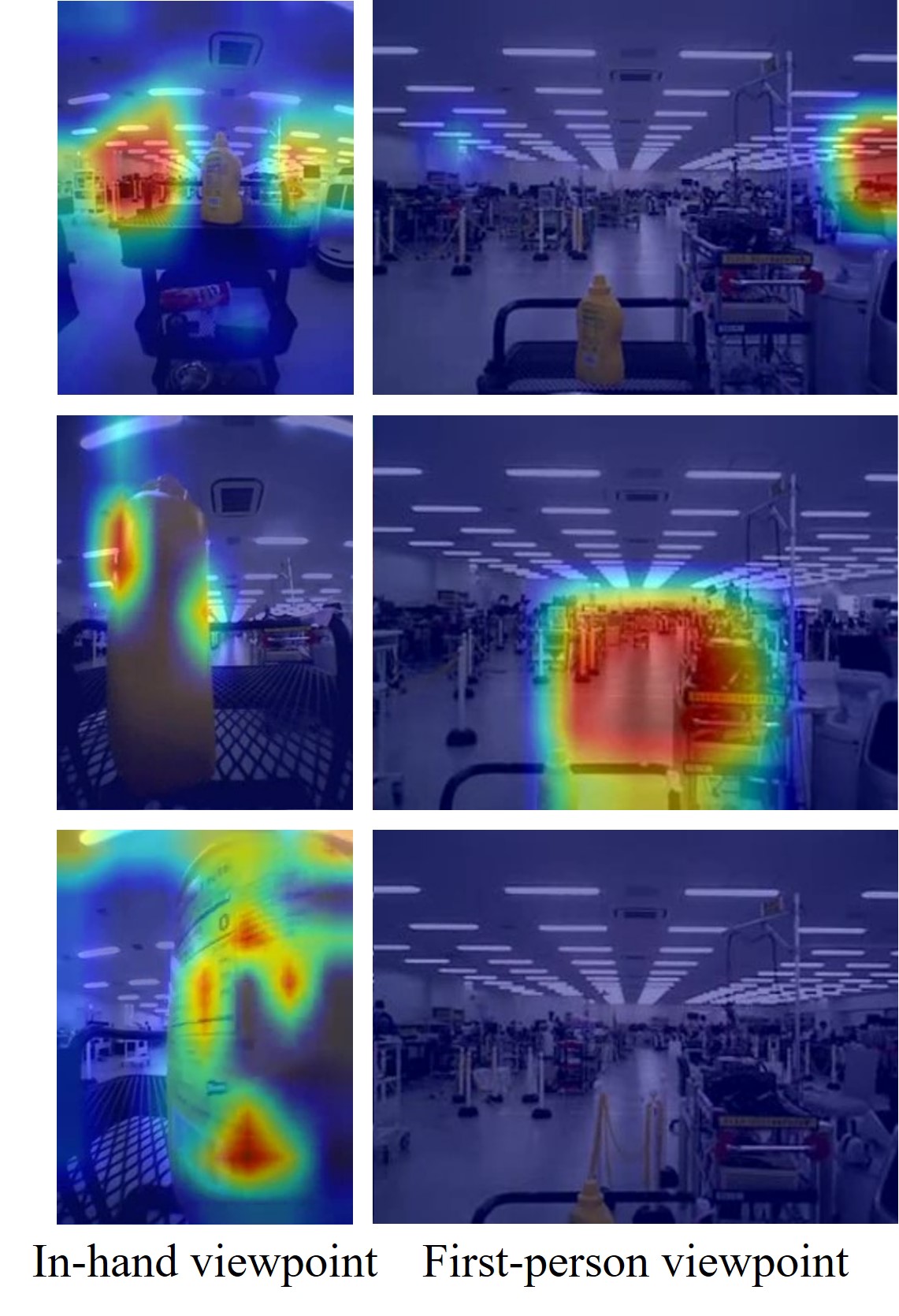}
        \caption{Visualize pixel attribution on the prediction of following methods in pick-bottle-from-shelf task. Top: only attention mechanism. Middle: only augmentation. Bottom: both attention mechanism and augmentation.}
        \label{fig:attention_pick_bottle_from_shelf}
        \vspace{-2mm}
    \end{center}
\end{figure}

\subsection{How can MM focus on task-related regions?}
\label{sec:experiments_region}


To clarify how the policy focused on task-related regions, the ablation study in the pick-bottle-from-shelf task is shown in Table \ref{tab:region}.
All methods are based on our proposed method described in section \ref{sec:experiments_impl}, and three types are compared: only attention mechanism is used, only augmentation is used, and both attention mechanism and augmentation are used.
We visualized the pixel attribution on the prediction using GradCAM.

In the case where only attention mechanism was used, the policy was unable to focus on task-related regions (bottle) as shown in Fig. \ref{fig:attention_pick_bottle_from_shelf} (top), and decreased the average success rate of the three environments by 33.3 points.
In the case of only augmentation, the policy was able to focus on task-related regions with \(o_h\) as shown in Fig. \ref{fig:attention_pick_bottle_from_shelf} (left of middle).
However, the policy was unable to focus on task-related regions with \(o_f\) as shown in Fig. \ref{fig:attention_pick_bottle_from_shelf} (right of middle).
The attribution was frozen during task execution as shown in Fig. \ref{fig:attention_pick_bottle_from_shelf} (right of middle), which showed that the policy could not learn task-related regions correctly.
The policy decreased the average success rate of the three environments by 17 points.
In the case of both attention mechanism and augmentation, the policy was able to focus on task-related regions as shown in Fig. \ref{fig:attention_pick_bottle_from_shelf} (bottom), and resulted in the best success rate under domain shift.
Thus, the result indicates the importance of ensuring a mechanism that focuses on task-related regions with attention mechanism and facilitates the learning of that with augmentation.

\section{Conclusions}
\label{sec:conclusions}

We propose a robust imitation learning method for MMs that focuses on task-related viewpoints and spatial regions when observing multiple viewpoints.
The multiple viewpoint policy includes attention mechanism, which is learned with an augmented dataset, and brings optimal viewpoints and robust visual embedding against occlusion and domain shift.


In the future, the model should be validated in dynamic and more complex environments.
For this purpose, the model should be extended to use other observation modalities for considering cases in which changes cannot be observed visually.
Its robustness to changes in task-related regions, such as changes in bottle color, should be enhanced.
Additionally, its robustness should extend beyond changes in observations to encompass changes in actions as well.






\bibliographystyle{IEEEtran}
\bibliography{bibtex/bib/reference}

\end{document}